\documentclass{article}
\usepackage{ijcai15}

\usepackage{amsmath}
\usepackage{amsfonts}
\usepackage{booktabs}
\usepackage[pdftex]{graphicx}
\usepackage{times}
\title{Imaging Time-Series to Improve Classification and Imputation}
\author{Zhiguang Wang \and Tim Oates\\
Department of Computer Science and Electric Engineering\\
University of Maryland, Baltimore County \\
\{stephen.wang, oates\}@umbc.edu}

\begin{document}

\maketitle
\begin{abstract} 
Inspired by recent successes of deep learning in computer vision, we
propose a novel framework for encoding time series as different types
of images, namely, Gramian Angular Summation/Difference Fields
(GASF/GADF) and Markov Transition Fields (MTF). This enables the use
of techniques from computer vision for time series classification and
imputation. We used Tiled Convolutional Neural Networks (tiled CNNs)
on 20 standard datasets to learn high-level features from the
individual and compound GASF-GADF-MTF images. Our approaches achieve
highly competitive results when compared to nine of the current best
time series classification approaches. Inspired by the bijection
property of GASF on 0/1 rescaled data, we train Denoised Auto-encoders
(DA) on the GASF images of four standard and one synthesized compound
dataset. The imputation MSE on test data is reduced by 12.18\%-48.02\%
when compared to using the raw data. An analysis of the features and
weights learned via tiled CNNs and DAs explains why the approaches
work.
\end{abstract}

\section{Introduction}
Since 2006, the techniques developed from deep neural networks (or, deep learning) have greatly impacted natural language processing, speech recognition and  computer vision research \cite{bengio2009learning,LiDeep2014}. One successful deep learning architecture used in computer vision is convolutional neural networks (CNN) \cite{lecun1998gradient}. CNNs
exploit translational invariance by extracting
features through receptive fields \cite{hubel1962receptive} and learning
with weight sharing, becoming the state-of-the-art approach in various
image recognition and computer vision tasks
\cite{krizhevsky2012imagenet}.  
Since
unsupervised pretraining has been shown to improve performance
\cite{erhan2010does}, sparse coding and Topographic Independent
Component Analysis (TICA) are integrated as unsupervised pretraining
approaches to learn more diverse features with complex invariances
\cite{kavukcuoglu2010learning,ngiam2010tiled}.

Along with the success of unsupervised pretraining applied in deep learning, others are studying unsupervised learning algorithms for generative models, such as Deep Belief Networks (DBN) and Denoised Auto-encoders (DA) \cite{hinton2006fast,vincent2008extracting}.  Many deep generative models are developed based on energy-based model or auto-encoders. Temporal autoencoding is integrated with Restrict Boltzmann Machines (RBMs) to improve generative models \cite{hausler2013temporal}. A training strategy inspired by recent work on optimization-based learning is proposed to train complex neural networks for imputation tasks \cite{brakel2013training}. A generalized Denoised Auto-encoder extends the theoretical framework and is applied to Deep Generative Stochastic Networks (DGSN) \cite{bengio2013generalized,bengio2013deep}.

Inspired by recent successes of supervised and unsupervised learning techniques in computer vision, we consider the problem of encoding time series as images to
allow machines to ``visually'' recognize, classify and learn structures and patterns. Reformulating features of time series as visual clues has raised much attention in computer science and physics. In speech recognition systems,
acoustic/speech data input is typically represented by
concatenating Mel-frequency cepstral coefficients (MFCCs) or
perceptual linear predictive coefficient (PLPs)
\cite{hermansky1990perceptual}. Recently, researchers are trying to build different network structures from time series for visual inspection or designing distance measures. Recurrence Networks were proposed to analyze the structural properties of time series from complex systems \cite{donner2010recurrence,donner2011recurrence}. They build adjacency matrices from the predefined recurrence functions to interpret the time series as complex networks. \citeauthor{silva2013time} extended the recurrence plot paradigm for time series classification using compression distance \cite{silva2013time}. Another way to build a weighted adjacency matrix is extracting transition dynamics from the first order Markov matrix \cite{campanharo2011duality}. Although these maps demonstrate distinct topological properties among different time series, it remains unclear how these topological properties relate to the original time series since they have no exact inverse operations.

We present three novel representations for encoding time
series as images that we call the Gramian Angular Summation/Difference Field (GASF/GADF) and the
Markov Transition Field (MTF). We applied deep Tiled Convolutional
Neural Networks (Tiled CNN) \cite{ngiam2010tiled} to
 classify  time series images on 20 standard datasets. Our experimental results demonstrate 
our approaches achieve the best performance on 9 of 20 standard dataset compared with 9 previous and current best classification methods. Inspired by the bijection property of GASF on $0/1$ rescaled data, we train the Denoised Auto-encoder (DA) on the GASF images of 4 standard and a synthesized compound dataset. The imputation MSE on test data is reduced by 12.18\%-48.02\% compared to using the raw data.  An analysis of the features and
weights learned via tiled CNNs and DA explains why the approaches work. 


\section{Imaging Time Series}

We first introduce our two frameworks for encoding time series as
images. The first type of image is a Gramian Angular Field (GAF), in
which we represent time series in a polar coordinate system instead of
the typical Cartesian coordinates.  In the Gramian matrix, each
element is actually the cosine of the summation of angles. Inspired by
previous work on the duality between time series and complex networks
\cite{campanharo2011duality}, the main idea of the second framework,
the Markov Transition Field (MTF), is to build the Markov matrix of
quantile bins after discretization and encode the dynamic transition
probability in a quasi-Gramian matrix.

\subsection{Gramian Angular Field}
\begin{figure}[t!]
    \centering
    \includegraphics[scale = 0.23]{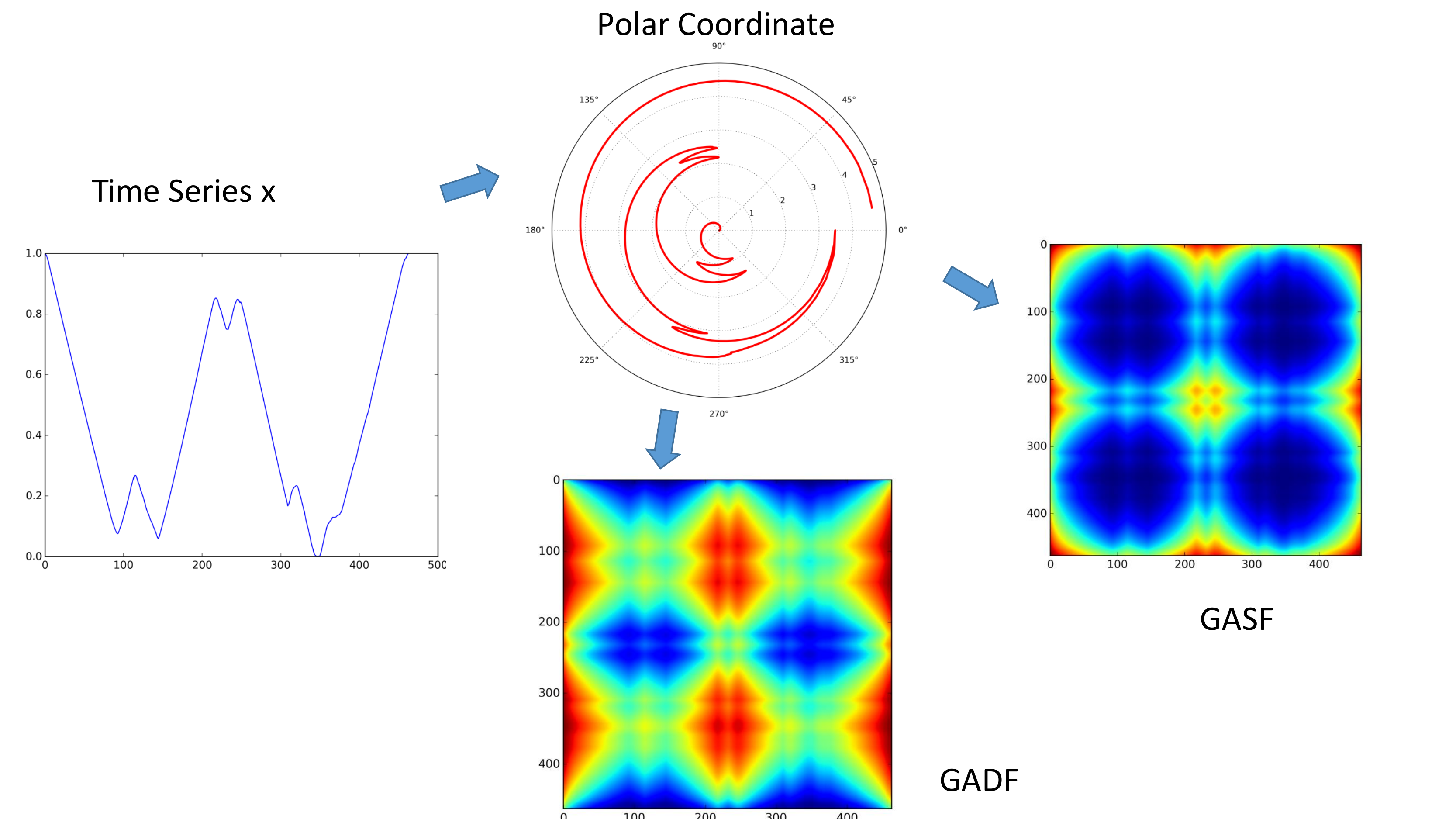}
    \caption{Illustration of the proposed encoding map of Gramian Angular Fields. $X$ is a sequence of rescaled time series in the 'Fish' dataset. We transform $X$ into a polar coordinate system by eq. (\ref{eqn:polar}) and finally calculate its GASF/GADF images with eqs. (\ref{eqn:GASF}) and (\ref{eqn:GADF}). In this example, we build GAFs without PAA smoothing, so the GAFs both have high resolution. } 
    \label{fig:encodingTS2GAF}
\end{figure}

Given a time series $X = \{x_1, x_2, ..., x_n\}$ of $n$ real-valued
observations, we rescale $X$ so that all values fall in the interval
$[-1,1]$ or $[0,1]$ by:
\begin{eqnarray}
& \tilde{x}_{-1}^i = \frac{(x_i-max(X)+(x_i-min(X))}{max(X)-min(X)}
\label{eqn:rescale-1} \\
\text{or} &\tilde{x}_{0}^i = \frac{x_i-min(X)}{max(X)-min(X)}
\label{eqn:rescale0}
\end{eqnarray}
Thus we can represent the rescaled time series
$\tilde{X}$ in polar coordinates by encoding the value as the
angular cosine and the time stamp as the radius with the equation below: 
\begin{eqnarray}
\left\{\begin{matrix}
\phi=\arccos{(\tilde{x_i})}, -1\leq \tilde{x_i} \leq 1, \tilde{x_i} \in \tilde{X}
\\r= \frac{t_i}{N}, t_i \in \mathbb{N}
\end{matrix}\right.
\label{eqn:polar}
\end{eqnarray}

In the equation above, $t_i$ is the time stamp and $N$ is a constant
factor to regularize the span of the polar coordinate system. This
polar coordinate based representation is a novel way to understand
time series. As time increases, corresponding values warp among
different angular points on the spanning circles, like water
rippling. The encoding map of equation \ref{eqn:polar} has two
important properties. First, it is bijective as $\cos(\phi)$ is
monotonic when $\phi \in [0,\pi]$. Given a time series, the proposed
map produces one and only one result in the polar coordinate system
with a unique inverse map. Second, as opposed to Cartesian
coordinates, polar coordinates preserve absolute temporal
relations. We will discuss this in more detail in future work.

Rescaled data in different intervals have different angular bounds. $[0,1]$ corresponds to the cosine function in $[0, \frac{\pi}{2}]$, while cosine values in the interval $[-1,1]$ fall into the angular bounds $[0, \pi]$. As we will discuss later, they provide different information granularity in the Gramian Angular Field for classification tasks, and the Gramian Angular Difference Field (GADF) of $[0,1]$ rescaled data has the accurate inverse map. This property actually lays the foundation for imputing missing value of time series by recovering the images.
   
After transforming the rescaled time series into the polar coordinate
system, we can easily exploit the angular perspective by considering
the trigonometric sum/difference between each point to identify the temporal
correlation within different time intervals. The Gramian Summation Angular Field (GASF) and Gramian Difference Angular Field (GADF) are defined as
follows: 

\begin{eqnarray}
GASF   
&=& \begin{bmatrix}
\cos(\phi_i+\phi_j)
\end{bmatrix} 
\\
&=& \tilde{X}' \cdot \tilde{X} - \sqrt{I-\tilde{X}^2}' \cdot \sqrt{I-\tilde{X}^2} 
\label{eqn:GASF} 
\end{eqnarray}
\begin{eqnarray}
GADF   
&=& \begin{bmatrix}
\sin(\phi_i-\phi_j)  
\end{bmatrix} 
\\
&=& \sqrt{I-\tilde{X}^2}' \cdot \tilde{X} - \tilde{X}' \cdot \sqrt{I-\tilde{X}^2}  
\label{eqn:GADF} 
\end{eqnarray}

$I$ is the unit row vector $[1,1,...,1]$. After transforming to the polar
coordinate system, we take time series at each time step as a 1-D
metric space. By defining the inner product $<x,y> = x\cdot
y-\sqrt{1-x^2} \cdot \sqrt{1-y^2}$ and $<x,y> = \sqrt{1-x^2} \cdot
y- x \cdot \sqrt{1-y^2}$, two types of Gramian Angular Fields (GAFs) are actually quasi-Gramian matrices $[<\tilde{x_1},\tilde{x_1}>]$. \footnote{'quasi' because the functions $<x,y>$ we defined do not satisfy the property of linearity in inner-product space.}



The GAFs have several advantages. First, they provide a way to preserve  temporal dependency, since time increases as the position moves
from top-left to bottom-right. The GAFs contain temporal correlations
because $G_{(i,j||i-j|=k)}$ represents the relative correlation by
superposition/difference of directions with respect to time interval $k$. The main diagonal $G_{i,i}$ is the special case when $k=0$, which contains
the original value/angular information. From the main diagonal, we
can reconstruct the time series from the high level
features learned by the deep neural network. However, the GAFs are large
because the size of the Gramian matrix is $n\times n$ when the length of
the raw time series is $n$. To reduce the size of the GAFs, we apply
Piecewise Aggregation Approximation (PAA) \cite{keogh2000scaling} to smooth
the time series while preserving the trends. The full pipeline for
generating the GAFs is illustrated in Figure \ref{fig:encodingTS2GAF}.

\subsection{Markov Transition Field}

\begin{figure}[t]
    \centering
    \includegraphics[scale = 0.23]{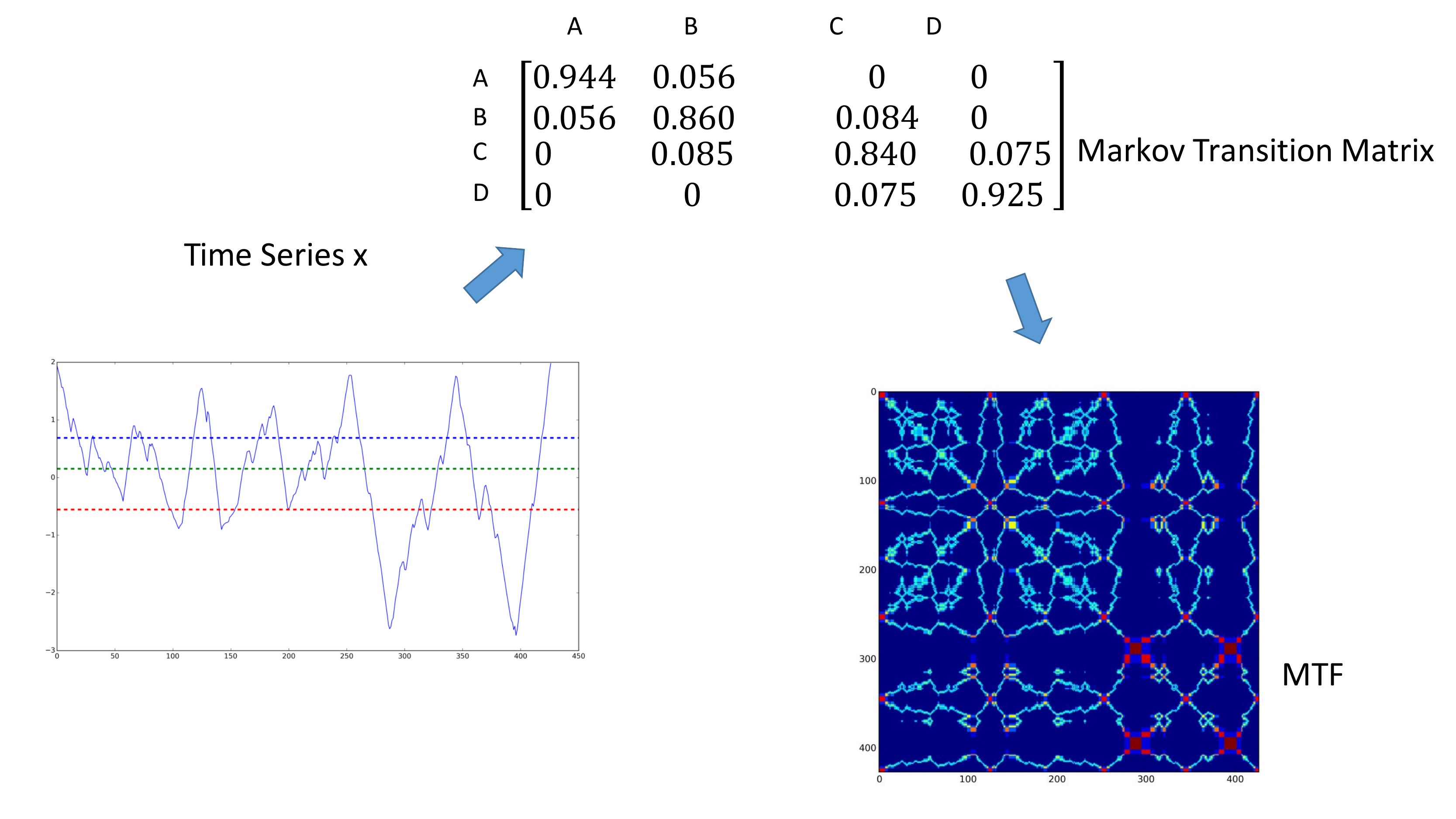}
    \caption{Illustration of the proposed encoding map of Markov Transition Fields. $X$ is a sequence of time-series in the 'ECG' dataset . $X$ is first discretized into $Q$ quantile bins. Then we calculate its Markov Transition Matrix $W$ and finally build its MTF with eq. (\ref{eqn:MTF}).}
    \label{fig:encodingTS2MTF}
\end{figure}

We propose a framework similar to \citeauthor{campanharo2011duality} for
encoding dynamical transition statistics, but we extend that idea by
representing the Markov transition probabilities sequentially to
preserve information in the time domain.  	

Given a time series $X$, we identify its $Q$ quantile bins and assign
each $x_i$ to the corresponding bins $q_j$ ($j \in [1,Q]$). Thus we
construct a $Q \times Q$ weighted adjacency matrix $W$ by counting
transitions among quantile bins in the manner of a first-order
Markov chain along the time axis. $w_{i,j}$ is given by the frequency
with which a point in quantile $q_j$ is followed by a point in quantile
$q_i$. After normalization by $\sum_j{w_{ij}=1}$, $W$ is the Markov
transition matrix.  It is insensitive to the distribution of $X$ and temporal dependency on time steps $t_i$. However, our experimental results on $W$ demonstrate that getting rid of the temporal dependency results in too much information loss in
matrix $W$. To overcome this drawback, we define the Markov Transition
Field (MTF) as follows:
\begin{equation}
M = 
\begin{bmatrix}
w_{ij|x_1 \in q_i,x_1 \in q_j}  & \cdots & w_{ij|x_1 \in q_i,x_n \in q_j} \\
w_{ij|x_2 \in q_i,x_1 \in q_j}  & \cdots & w_{ij|x_2 \in q_i,x_n \in q_j} \\
\vdots  & \ddots & \vdots \\
w_{ij|x_n \in q_i,x_1 \in q_j}  & \cdots & w_{ij|x_n \in q_i,x_n \in q_j} \\
\end{bmatrix} 
\label{eqn:MTF}
\end{equation} 

We build a $Q \times Q$ Markov transition matrix ($W$) by
dividing the data (magnitude) into $Q$ quantile bins. The quantile
bins that contain the data at time stamp $i$ and $j$ (temporal axis)
are $q_{i}$ and $q_{j}$ ($q \in [1,Q]$). $M_{ij}$ in the MTF denotes the
transition probability of $q_{i} \rightarrow q_{j}$. That is, we
spread out matrix $W$ which contains the transition probability on the
magnitude axis into the MTF matrix by considering the temporal positions.

By assigning the probability from the quantile at time step $i$ to the
quantile at time step $j$ at each pixel $M_{ij}$, the MTF $M$ actually
encodes the multi-span transition probabilities of the time
series. $M_{i,j||i-j|=k}$ denotes the transition probability between
the points with time interval $k$. For example, $M_{ij|j-i=1}$
illustrates the transition process along the time axis with a skip
step. The main diagonal $M_{ii}$, which is a special case when $k=0$
captures the probability from each quantile to itself (the
self-transition probability) at time step $i$. To make the image size
manageable and computation more efficient, we reduce the MTF size by
averaging the pixels in each non-overlapping $m \times m$ patch with
the blurring kernel $\{\frac{1}{m^2}\}_{m \times m}$. That is, we
aggregate the transition probabilities in each subsequence of length
$m$ together. Figure \ref{fig:encodingTS2MTF} shows the procedure to
encode time series to MTF. 

\begin{figure}[t]
    \centering
    \includegraphics[scale = 0.23]{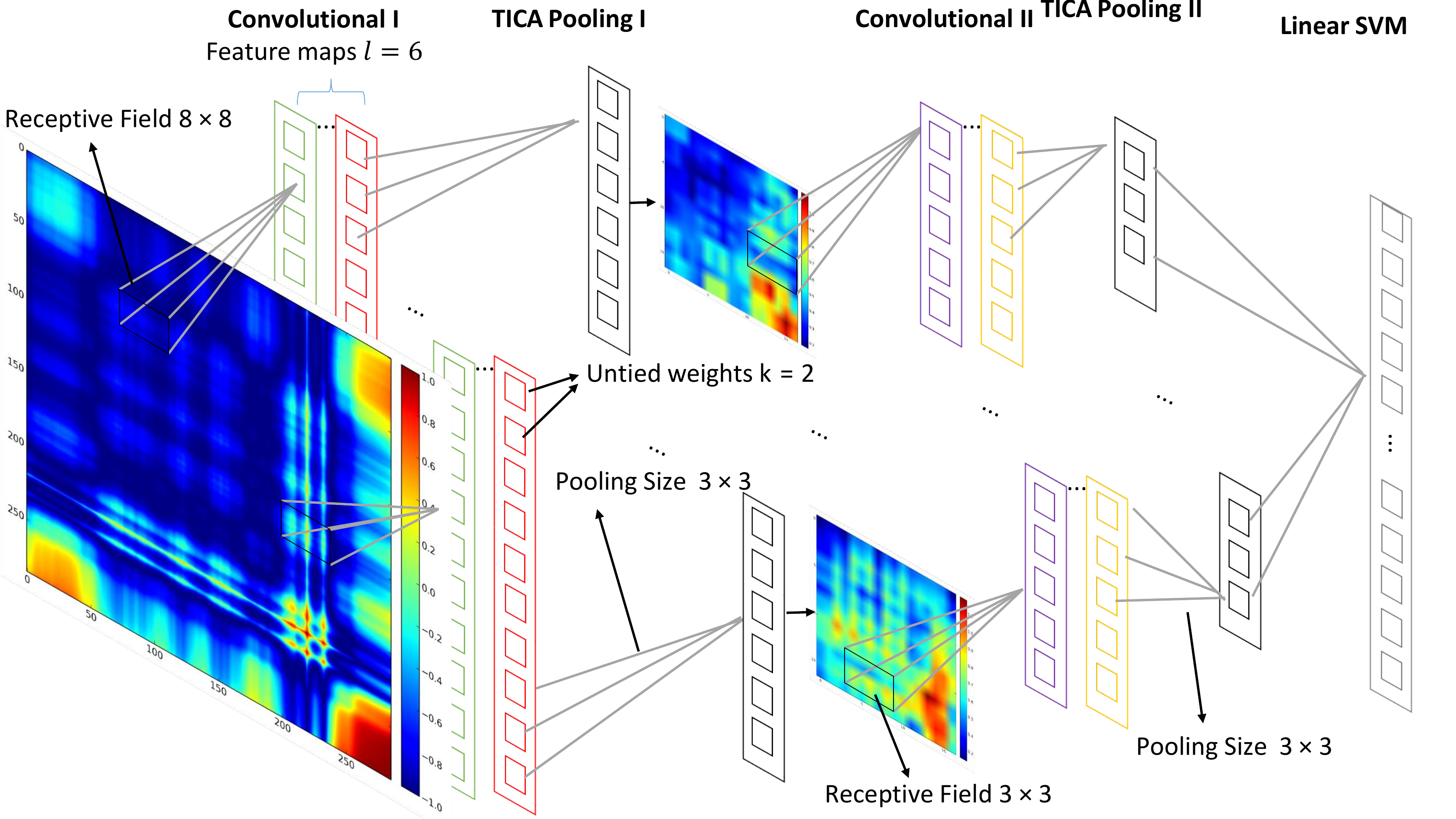}
    \caption{Structure of the tiled convolutional neural networks. We
      fix the size of receptive fields to $8 \times 8$ in the first
      convolutional layer and $3 \times 3$ in the second convolutional
      layer. Each TICA pooling layer pools over a block of $3 \times
      3$ input units in the previous layer without warping around the borders to optimize for sparsity of the pooling units. The number of pooling units in each map is exactly the same as the number of input units. The last layer is a linear SVM for classification. We construct this network by stacking two Tiled CNNs, each with 6 maps ($l=6$) and tiling size $k={1,2,3}$.}
    \label{fig:TCNN_structure}
\end{figure}

\begin{table*}[t]
\small
  \centering
  \caption{Summary of error rates for 3 classic baselines, 6 recently published best results and our approach. The symbols $\triangleleft$, $\ast$, $\dagger$ and $\bullet$ represent datasets generated from human motions, figure shapes, synthetically predefined procedures and all remaining temporal signals, respectively. For our approach, the numbers in brackets are the optimal PAA size and quantile size.}
    \begin{tabular}{rrrrrrrrrrr}
    \toprule
    Dataset & 1NN- & 1NN-DTW- & 1NN-DTW- & Fast- & SAX- & SAX- & RPCD & SMTS & TSBF & GASF-GADF- \\
      & RAW & BWW & nWW & Shapelet & BoP & VSM &   &   &   & MTF \\
    \midrule
    50words $\bullet$  & 0.369 & 0.242 & 0.31 & N/A & 0.466 & N/A & 0.2264 & 0.289 & \textbf{0.209} & 0.301 (16, 32) \\
    Adiac $\ast$ & 0.389 & 0.391 & 0.396 & 0.514 & 0.432 & 0.381 & 0.3836 & 0.248 & \textbf{0.245} & 0.373 (32, 48) \\
    Beef $\bullet$ & 0.467 & 0.467 & 0.5 & 0.447 & 0.433 & 0.33 & 0.3667 & 0.26 & 0.287 & \textbf{0.233 (64, 40)} \\
    CBF $\dagger$ & 0.148 & 0.004 & 0.003 & 0.053 & 0.013 & 0.02 & N/A & 0.02 & \textbf{0.009} & \textbf{0.009 (32, 24)} \\
    Coffee $\bullet$ & 0.25 & 0.179 & 0.179 & 0.068 & 0.036 & \textbf{0} & \textbf{0} & 0.029 & 0.004 & \textbf{0 (64, 48)} \\
    ECG $\bullet$ & 0.12 & 0.12 & 0.23 & 0.227 & 0.15 & 0.14 & 0.14 & 0.159 & 0.145 & \textbf{0.09 (8, 32)} \\
    FaceAll $\ast$ & 0.286 & 0.192 & 0.192 & 0.411 & 0.219 & 0.207 & \textbf{0.1905} & 0.191 & 0.234 & 0.237 (8, 48) \\
    FaceFour $\ast$ & 0.216 & 0.114 & 0.17 & 0.090 & 0.023 & \textbf{0} & 0.0568 & 0.165 & 0.051 & 0.068 (8, 16) \\
    fish $\ast$ & 0.217 & 0.16 & 0.167 & 0.197 & 0.074 & \textbf{0.017} & 0.1257 & 0.147 & 0.08 & 0.114 (8, 40) \\
    Gun\_Point $\triangleleft$ & 0.087 & 0.087 & 0.093 & 0.061 & 0.027 & 0.007 & \textbf{0} & 0.011 & 0.011 & 0.08 (32, 32) \\
    Lighting2 $\bullet$ & 0.246 & 0.131 & 0.131 & 0.295 & 0.164 & 0.196 & 0.2459 & 0.269 & 0.257 & \textbf{0.114 (16, 40)} \\
    Lighting7 $\bullet$ & 0.425 & 0.288 & 0.274 & 0.403 & 0.466 & 0.301 & 0.3562 & \textbf{0.255} & 0.262 & 0.260 (16, 48) \\
    OliveOil $\bullet$ & 0.133 & 0.167 & 0.133 & 0.213 & 0.133 & 0.1 & 0.1667 & 0.177 & \textbf{0.09} & 0.2 (8, 48) \\
    OSULeaf $\ast$ & 0.483 & 0.384 & 0.409 & 0.359 & 0.256 & \textbf{0.107} & 0.3554 & 0.377 & 0.329 & 0.358 (16, 32) \\
    SwedishLeaf $\ast$ & 0.213 & 0.157 & 0.21 & 0.269 & 0.198 & 0.01 & 0.0976 & 0.08 & 0.075 & \textbf{0.065 (16, 48)} \\
    synthetic control $\dagger$ & 0.12 & 0.017 & \textbf{0.007} & 0.081 & 0.037 & 0.251 & N/A & 0.025 & 0.008 & \textbf{0.007 (64, 48)} \\
    Trace $\dagger$ & 0.24 & 0.01 & \textbf{0} & 0.002 & \textbf{0} & \textbf{0} & N/A & \textbf{0} & 0.02 & \textbf{0 (64, 48)} \\
    Two Patterns $\dagger$ & 0.09 & 0.0015 & \textbf{0} & 0.113 & 0.129 & 0.004 & N/A & 0.003 & 0.001 & 0.091 (64, 32) \\
    wafer $\bullet$ & 0.005 & 0.005 & 0.02 & 0.004 & 0.003 & 0.0006 & 0.0034 & \textbf{0} & 0.004 & \textbf{0 (64, 16)} \\
    yoga $\ast$ & 0.17 & 0.155 & 0.164 & 0.249 & 0.17 & 0.164 & 0.134 & \textbf{0.094} & 0.149 & 0.196 (8, 32) \\
    \textbf{\#wins} & 0 & 0 & 3 & 0 & 1 & 5 & 3 & 4 & 4 & \textbf{9} \\
    \bottomrule
    \end{tabular}%
  \label{tab:fullstatistics}%
\end{table*}%

\section{Classify Time Series Using GAF/MTF with Tiled CNNs}

We apply Tiled CNNs to classify time series using GAF and MTF representations on 20 datasets from \cite{keogh2011ucr} in different domains such as medicine, entomology, engineering, astronomy, signal processing, and others. The datasets are pre-split into training and testing sets to facilitate experimental comparisons. We compare the classification error rate of our GASF-GADF-MTF approach with previously published results of 3 competing methods and 6 best approaches proposed recently: early state-of-the-art 1NN classifiers based on Euclidean distance and DTW (Best Warping Window and no Warping Window), Fast-Shapelets\cite{rakthanmanon2013fast}, a 1NN classifier based on SAX with Bag-of-Patterns (SAX-BoP) \cite{lin2012rotation}, a SAX based Vector Space Model (SAX-VSM)\cite{senin2013sax}, a classifier based on the Recurrence Patterns Compression Distance(RPCD) \cite{silva2013time}, a tree-based symbolic representation for multivariate time series (SMTS) \cite{baydogan2014learning} and a SVM classifier based on a bag-of-features representation (TSBF) \cite{baydogan2013bag}.

\subsection{Tiled Convolutional Neural Networks}
Tiled Convolutional Neural Networks are a
variation of Convolutional Neural Networks that use tiles and multiple
feature maps to learn invariant features. Tiles are parameterized by a
tile size $k$ to control the distance over which weights are
shared. By producing multiple feature maps, Tiled CNNs learn
overcomplete representations through unsupervised pretraining with
Topographic ICA (TICA). For the sake of space, please refer to \cite{ngiam2010tiled} for more details. The structure of Tiled CNNs applied in this paper is illustrated in Figure \ref{fig:TCNN_structure}.

\subsection{Experiment Setting}

In our experiments, the size of the GAF image is regulated by the the
number of PAA bins $S_{GAF}$. Given a time series $X$ of size $n$, we
divide the time series into $S_{GAF}$ adjacent, non-overlapping
windows along the time axis and extract the means of each bin. This
enables us to construct the smaller GAF matrix $G_{S_{GAF} \times
  S_{GAF}}$. MTF requires the time series to be discretized into $Q$
quantile bins to calculate the $Q \times Q$ Markov transition matrix,
from which we construct the raw MTF image $M_{n \times n}$
afterwards. Before classification, we shrink the MTF image size to
$S_{MTF} \times S_{MTF}$ by the blurring kernel $\{\frac{1}{m^2}\}_{m
\times  m}$ where $m=\lceil \frac{n}{S_{MTF}}\rceil$. The Tiled CNN
is trained with image size $\{S_{GAF},S_{MTF}\} \in \{16, 24, 32, 40,
48\}$ and quantile size $Q \in \{8, 16, 32, 64\}$. At the last layer
of the Tiled CNN, we use a linear soft margin SVM
\cite{fan2008liblinear} and select $C$ by 5-fold cross validation
over $\{10^{-4}, 10^{-3}, \ldots, 10^4\}$ on the training set. 

For each input of image size $S_{GAF}$ or $S_{MTF}$ and quantile size
$Q$, we pretrain the Tiled CNN with the full unlabeled dataset (both training and test set) to learn
the initial weights $W$ through TICA. Then we train the SVM at the
last layer by selecting the penalty factor $C$ with cross
validation. Finally, we classify the test set using the optimal
hyperparameters $\{S, Q, C\}$ with the lowest error rate on the
training set. If two or more models tie, we prefer the larger $S$
and $Q$ because larger $S$ helps preserve more information through
the PAA procedure and larger $Q$ encodes the dynamic transition
statistics with more detail. Our model selection approach provides
generalization without being overly expensive computationally.

\subsection{Results and Discussion}

We use Tiled CNNs to classify the single GASF, GADF and MTF images as well as the compound GASF-GADF-MTF images
on 20 datasets. For the sake of space, we do not show the full results on single-channel images. Generally, our approach is not prone to
overfitting by the relatively small difference between
training and test set errors.  One exception is the Olive Oil dataset
with the MTF approach where the test error is significantly higher.

In addition to the risk of potential overfitting, we found that MTF has generally
higher error rates than GAFs. This is most likely because of
the uncertainty in the inverse map of MTF. Note that the encoding
function from $-1/1$ rescaled time series to GAFs and MTF are both surjections. The map
functions of GAFs and MTF will each produce only one image with
fixed $S$ and $Q$ for each given time series $X$ . Because they are both surjective mapping functions, the inverse
image of both mapping functions is not fixed. However, the mapping function of GAFs on $0/1$ rescaled time series are bijective. As shown in a later
section, we can reconstruct the raw time series from
the diagonal of GASF, but it is very hard to even roughly recover the signal from
MTF. Even for $-1/1$ rescaled data, the GAFs have smaller uncertainty in the inverse image of their mapping
function because such randomness only comes from the ambiguity of
$\cos(\phi)$ when $\phi \in [0,2\pi]$. MTF, on the other hand, has a
much larger inverse image space, which results in large variations when
we try to recover the signal. Although MTF encodes the transition
dynamics which are important features of time series, such features
alone seem not to be sufficient for recognition/classification tasks.

Note that at each pixel, $G_{ij}$ denotes the superstition/difference of the
directions at $t_i$ and $t_j$, $M_{ij}$ is the transition probability
from the quantile at $t_i$ to the quantile at $t_j$. GAF encodes 
static information while MTF depicts information about dynamics. From this
point of view, we consider them as three ``orthogonal'' channels, like
different colors in the RGB image space.  Thus, we can combine GAFs and MTF
images of the same size (i.e. $S_{GAFs}=S_{MTF}$) to construct a
triple-channel image (GASF-GADF-MTF). It combines both the static
and dynamic statistics embedded in the raw time series, and we posit that it
will be able to enhance classification performance. In the
experiments below, we pretrain and tune the Tiled CNN on the compound
GASF-GADF-MTF images. Then, we report the classification error rate on test
sets. In Table
\ref{tab:fullstatistics}, the Tiled CNN classifiers on GASF-GADF-MTF images achieved significantly competitive results with 9 other state-of-the-art time series classification approaches.  

\section{Image Recovery on GASF for Time Series Imputation with Denoised Auto-encoder}

\begin{figure}[t]
    \centering
    \includegraphics[scale = 0.27]{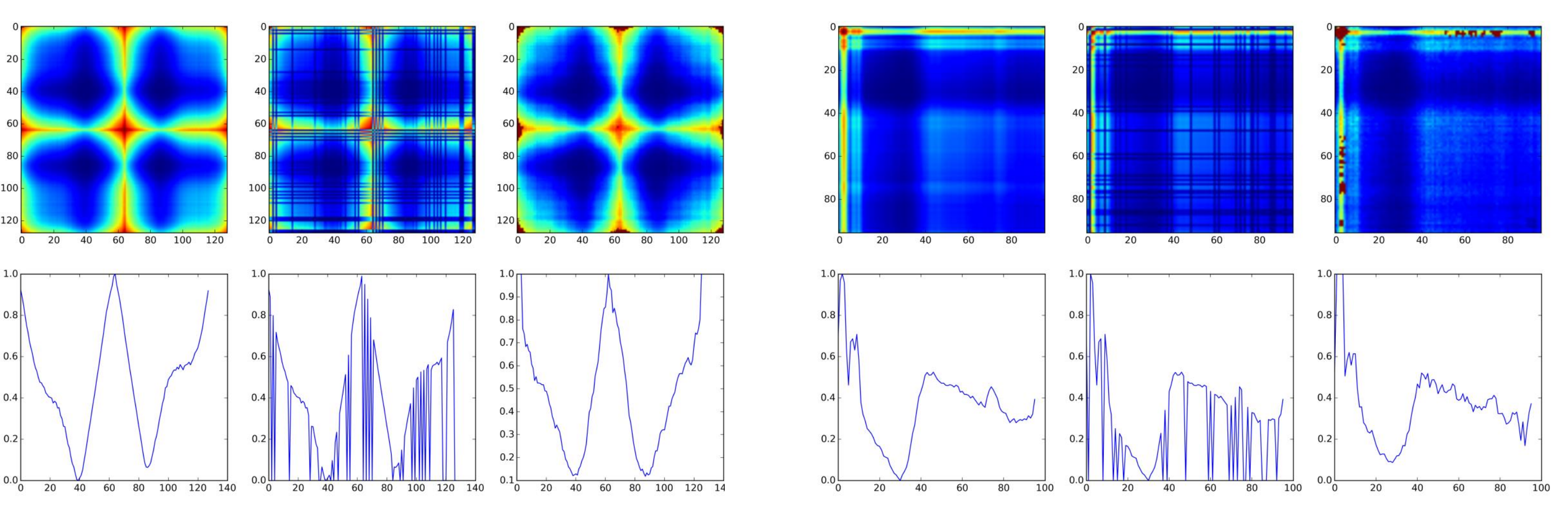}
    \caption{Pipeline of time series imputation by image recovery. Raw GASF $\rightarrow$ "broken" GASF $\rightarrow$ recovered GASF (top), Raw time series $\rightarrow$ corrupted time series with missing value $\rightarrow$ predicted time series (bottom) on dataset "SwedishLeaf" (left) and "ECG" (right).}
    \label{fig:impute_proc}
\end{figure}

As previously mentioned, the mapping functions from $-1/1$ rescaled time series to GAFs are
surjections. The uncertainty among the inverse images come from the
ambiguity of the $\cos(\phi)$  when $\phi \in [0,2\pi]$. However the mapping functions of $0/1$ rescaled time series are bijections. The main
diagonal of GASF, i.e. $\{G_{ii}\} =\{\cos(2\phi_{i})\}$ allows us to
precisely reconstruct the original time series by  
\begin{eqnarray}
\cos(\phi) = \sqrt{\frac{\cos(2\phi)+1}{2}} && \phi \in [0, \frac{\pi}{2}]
\label{eqn:bijection}
\end{eqnarray}

Thus, we can predict missing values among time series through recovering the "broken" GASF images. During training, we manually add "salt-and-pepper" noise (i.e., randomly set a number of points to 0) to the raw time series and transform the data to GASF images. Then a single layer Denoised Auto-encoder (DA) is fully trained as a generative model to reconstruct GASF images. Note that at the input layer, we do not add noise again to the "broken" GASF images. A Sigmoid function helps to learn the nonlinear features at the hidden layer. At the last layer we compute the Mean Square Error (MSE) between the original and "broken"  GASF images as the loss function to evaluate fitting performance. To train the models simple batch gradient descent is applied to back propagate the inference loss. For testing, after we corrupt the time series and transform the noisy data to "broken" GASF, the trained DA helps recover the image, on which we extract the main diagonal to reconstruct the recovered time series. To compare the imputation performance, we also test standard DA with the raw time series data as input to recover the missing values (Figure. \ref{fig:impute_proc}).

\subsection{Experiment Setting}
For the DA models we use batch gradient descent with a batch size of 20. Optimization iterations run until the MSE changed less than a threshold of $10^{-3}$ for GASF and $10^{-5}$ for raw time series. A single hidden layer has 500 hidden neurons with sigmoid functions. We choose four dataset of different types from the UCR time series repository for the imputation task: "Gun Point" (human motion), "CBF" (synthetic data), "SwedishLeaf" (figure shapes) and "ECG" (other remaining temporal signals). To explore if the statistical dependency learned by the DA can be generalized to unknown data, we use the above four datasets and the  "Adiac" dataset together to train the DA to impute two totally unknown test datasets, "Two Patterns" and "wafer" (We name these synthetic miscellaneous datasets "7 Misc"). To add randomness to the input of DA, we randomly set $20\%$ of the raw data among a specific time series to be zero (salt-and-pepper noise). Our experiments for imputation are implemented with Theano \cite{Bastien-Theano-2012}. To control for the random initialization of the parameters and the randomness induced by  gradient descent, we repeated every experiment 10 times and report the average MSE.

\subsection{Results and Discussion}   
\begin{table}[h]
  \centering
  \caption{MSE of imputation on time series using raw data and GASF images. }
    \begin{tabular}{rrrrr}
    \toprule
    Dataset & \multicolumn{2}{c}{Full MSE} & \multicolumn{2}{c}{Interpolation MSE} \\
    \midrule
      & \multicolumn{1}{c}{Raw} & \multicolumn{1}{c}{GASF} & \multicolumn{1}{c}{Raw} & \multicolumn{1}{c}{GASF} \\
    ECG & 0.01001 & 0.01148 & 0.02301 & 0.01196 \\
    CBF & 0.02009 & 0.03520 & 0.04116 & 0.03119 \\
    Gun Point & 0.00693 & 0.00894 & 0.01069 & 0.00841 \\
    SwedishLeaf & 0.00606 & 0.00889 & 0.01117 & 0.00981 \\
    7 Misc & 0.06134 & 0.10130 & 0.10998 & 0.07077 \\
    \bottomrule
    \end{tabular}%
  \label{tab:impute_results}%
\end{table}%

In Table \ref{tab:impute_results}, "Full MSE" means the MSE between the complete recovered and original sequence and "Imputation MSE" means the MSE of only the unknown points among each time series. Interestingly, DA on the raw data perform well on the whole sequence, generally, but there is a gap between the full MSE and imputation MSE. That is, DA on raw time series can fit the known data much better than predicting the unknown data (like overfitting). Predicting the missing value using GASF always achieves slightly higher full MSE but the imputation MSE is reduced by 12.18\%-48.02\%. We can observe that the difference between the full MSE and imputation MSE is much smaller on GASF than on the raw data. Interpolation with GASF has more stable performance than on the raw data.

Why does predicting missing values using GASF have more stable performance than using raw time series? Actually, the transformation maps of GAFs are generally equivalent to a kernel trick. By defining the inner product $k(x_i,x_j)$, we achieve data augmentation by increasing the dimensionality of the raw data. By preserving the temporal and spatial information in GASF images, the DA utilizes both temporal and spatial dependencies by considering the missing points as well as their relations to other data that has been explicitly encoded in the GASF images. Because the entire sequence, instead of a short subsequence,  helps predict the missing value, the performance is more stable as the full MSE and imputation MSE are close.         

\section{Analysis on Features and Weights Learned by Tiled CNNs and DA}

\begin{figure}[t]
    \centering
    \includegraphics[scale = 0.25]{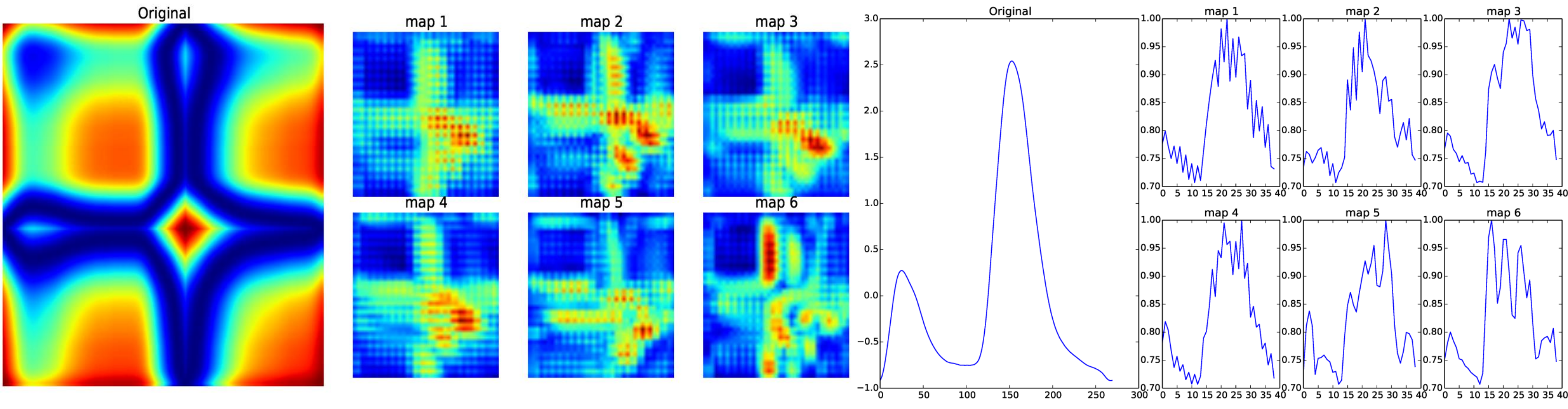}
    \caption{(a) Original GASF and its six learned feature maps before the SVM layer in Tiled CNNs (left). (b) Raw time series and its reconstructions from the main diagonal of six feature maps on '50Words' dataset (right).}
    \label{fig:feature_reconstruction}
\end{figure}

In contrast to the cases in which the CNNs is applied in natural image
recognition tasks, neither GAFs nor MTF have natural interpretations of
visual concepts like ``edges'' or ``angles''.  In this section we
analyze the features and weights learned through Tiled CNNs to explain
why our approach works.     


Figure \ref{fig:feature_reconstruction} illustrates the reconstruction
results from six feature maps learned through the Tiled CNNs on GASF (by Eqn \ref{eqn:bijection}). The Tiled CNNs extracts the color patch, which is essentially a 
moving average that enhances several receptive fields within the nonlinear
units by different trained weights. It is not a simple moving average but the synthetic
integration by considering the 2D temporal dependencies among
different time intervals, which is a benefit from the Gramian 
matrix structure that helps preserve the temporal information. By observing the orthogonal reconstruction from each layer of the feature maps, we can clearly observe that the tiled CNNs can extract the multi-frequency dependencies through the convolution and pooling architecture on the GAF and MTF images to preserve the trend while addressing more details in different subphases. The high-leveled feature maps
learned by the Tiled CNN are equivalent to a multi-frequency approximator
of the original curve. Our experiments also demonstrates the
learned weight matrix $W$ with the constraint $WW^T=I$, which
makes effective use of local orthogonality. The TICA pretraining
provides the built-in advantage that the function w.r.t the parameter
space is not likely to be ill-conditioned as $WW^T=1$. The weight matrix $W$ is quasi-orthogonal and approaching 0 without large
magnitude. This implies that the condition number of $W$ approaches 1 and  helps the system to be well-conditioned. 

\begin{figure}[t]
    \centering
    \includegraphics[scale = 0.25]{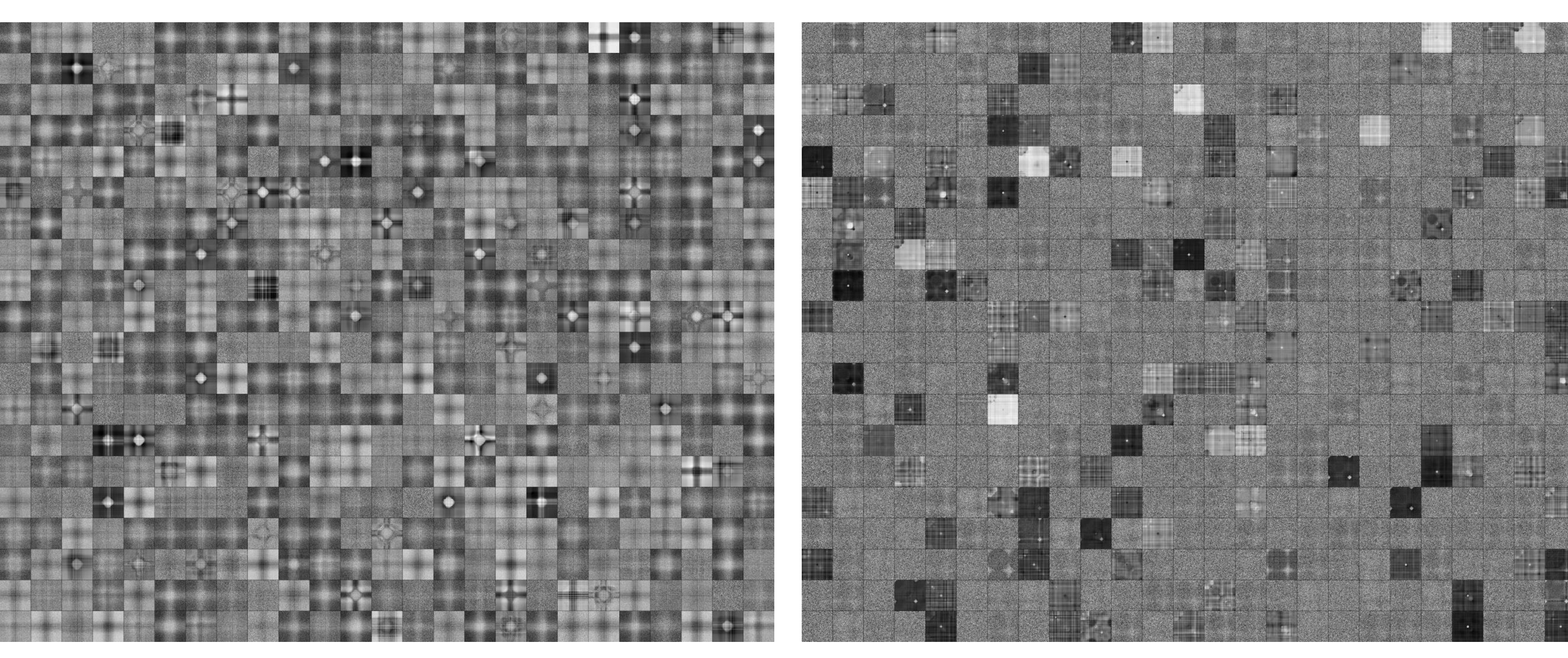}
    \caption{All 500 filters learned by DA on the "Gun Point" (left) and "7 Misc" (right) dataset.} 
    \label{fig:GP_featuremap_impute}
\end{figure}

As for imputation,  because the GASF images have no concept of "angle" and "edge", DA actually learned different prototypes of the GASF images (Table \ref{fig:GP_featuremap_impute}). We find that there is significant noise in the filters on the "7 Misc" dataset because the training set is relatively small to better learn different filters. Actually, all the noisy filters with no patterns work like one Gaussian noise filter.  

\section{Conclusions and Future Work}

We created a pipeline for converting time series into novel
representations, GASF, GADF and MTF images, and extracted multi-level features
from these using Tiled CNN and DA for classification and imputation. We demonstrated that our approach yields competitive
results for classification when compared to recently best methods. Imputation using GASF achieved better and more stable performance than on the raw data using DA. Our analysis of the
features learned from Tiled CNN suggested that Tiled CNN works like a
multi-frequency moving average that benefits from the 2D temporal
dependency that is preserved by Gramian matrix. Features learned by DA on GASF is shown to be different prototype, as correlated basis to construct the raw images.   

Important future work will involve developing recurrent neural nets
to process streaming data.  We are also quite interested in how different deep learning architectures perform on the GAFs and MTF images. Another important future work is to learn deep generative models with more high-level features on GAFs images. We aim to  further apply our time series models in real world regression/imputation and anomaly detection tasks. 
    
\newpage
\small
\bibliographystyle{named} 
\bibliography{shortbibfile}

\begin{thebibliography}{}

\bibitem[\protect\citeauthoryear{Bastien \bgroup \em et al.\egroup
  }{2012}]{Bastien-Theano-2012}
Fr{\'{e}}d{\'{e}}ric Bastien, Pascal Lamblin, Razvan Pascanu, James Bergstra,
  Ian~J. Goodfellow, Arnaud Bergeron, Nicolas Bouchard, and Yoshua Bengio.
\newblock Theano: new features and speed improvements.
\newblock Deep Learning and Unsupervised Feature Learning NIPS 2012 Workshop,
  2012.

\bibitem[\protect\citeauthoryear{Baydogan and
  Runger}{2014}]{baydogan2014learning}
Mustafa~Gokce Baydogan and George Runger.
\newblock Learning a symbolic representation for multivariate time series
  classification.
\newblock {\em Data Mining and Knowledge Discovery}, pages 1--23, 2014.

\bibitem[\protect\citeauthoryear{Baydogan \bgroup \em et al.\egroup
  }{2013}]{baydogan2013bag}
Mustafa~Gokce Baydogan, George Runger, and Eugene Tuv.
\newblock A bag-of-features framework to classify time series.
\newblock {\em Pattern Analysis and Machine Intelligence, IEEE Transactions
  on}, 35(11):2796--2802, 2013.

\bibitem[\protect\citeauthoryear{Bengio and
  Thibodeau-Laufer}{2013}]{bengio2013deep}
Yoshua Bengio and Eric Thibodeau-Laufer.
\newblock Deep generative stochastic networks trainable by backprop.
\newblock {\em arXiv preprint arXiv:1306.1091}, 2013.

\bibitem[\protect\citeauthoryear{Bengio \bgroup \em et al.\egroup
  }{2013}]{bengio2013generalized}
Yoshua Bengio, Li~Yao, Guillaume Alain, and Pascal Vincent.
\newblock Generalized denoising auto-encoders as generative models.
\newblock In {\em Advances in Neural Information Processing Systems}, pages
  899--907, 2013.

\bibitem[\protect\citeauthoryear{Bengio}{2009}]{bengio2009learning}
Yoshua Bengio.
\newblock Learning deep architectures for ai.
\newblock {\em Foundations and trends{\textregistered} in Machine Learning},
  2(1):1--127, 2009.

\bibitem[\protect\citeauthoryear{Brakel \bgroup \em et al.\egroup
  }{2013}]{brakel2013training}
Phil{\'e}mon Brakel, Dirk Stroobandt, and Benjamin Schrauwen.
\newblock Training energy-based models for time-series imputation.
\newblock {\em The Journal of Machine Learning Research}, 14(1):2771--2797,
  2013.

\bibitem[\protect\citeauthoryear{Campanharo \bgroup \em et al.\egroup
  }{2011}]{campanharo2011duality}
Andriana~SLO Campanharo, M~Irmak Sirer, R~Dean Malmgren, Fernando~M Ramos, and
  Lu{\'\i}s A~Nunes Amaral.
\newblock Duality between time series and networks.
\newblock {\em PloS one}, 6(8):e23378, 2011.

\bibitem[\protect\citeauthoryear{Deng and Yu}{2014}]{LiDeep2014}
Li~Deng and Dong Yu.
\newblock Deep learning: Methods and applications.
\newblock Technical Report MSR-TR-2014-21, January 2014.

\bibitem[\protect\citeauthoryear{Donner \bgroup \em et al.\egroup
  }{2010}]{donner2010recurrence}
Reik~V Donner, Yong Zou, Jonathan~F Donges, Norbert Marwan, and J{\"u}rgen
  Kurths.
\newblock Recurrence networks—a novel paradigm for nonlinear time series
  analysis.
\newblock {\em New Journal of Physics}, 12(3):033025, 2010.

\bibitem[\protect\citeauthoryear{Donner \bgroup \em et al.\egroup
  }{2011}]{donner2011recurrence}
Reik~V Donner, Michael Small, Jonathan~F Donges, Norbert Marwan, Yong Zou,
  Ruoxi Xiang, and J{\"u}rgen Kurths.
\newblock Recurrence-based time series analysis by means of complex network
  methods.
\newblock {\em International Journal of Bifurcation and Chaos},
  21(04):1019--1046, 2011.

\bibitem[\protect\citeauthoryear{Erhan \bgroup \em et al.\egroup
  }{2010}]{erhan2010does}
Dumitru Erhan, Yoshua Bengio, Aaron Courville, Pierre-Antoine Manzagol, Pascal
  Vincent, and Samy Bengio.
\newblock Why does unsupervised pre-training help deep learning?
\newblock {\em The Journal of Machine Learning Research}, 11:625--660, 2010.

\bibitem[\protect\citeauthoryear{Fan \bgroup \em et al.\egroup
  }{2008}]{fan2008liblinear}
Rong-En Fan, Kai-Wei Chang, Cho-Jui Hsieh, Xiang-Rui Wang, and Chih-Jen Lin.
\newblock Liblinear: A library for large linear classification.
\newblock {\em The Journal of Machine Learning Research}, 9:1871--1874, 2008.

\bibitem[\protect\citeauthoryear{H{\"a}usler \bgroup \em et al.\egroup
  }{2013}]{hausler2013temporal}
Chris H{\"a}usler, Alex Susemihl, Martin~P Nawrot, and Manfred Opper.
\newblock Temporal autoencoding improves generative models of time series.
\newblock {\em arXiv preprint arXiv:1309.3103}, 2013.

\bibitem[\protect\citeauthoryear{Hermansky}{1990}]{hermansky1990perceptual}
Hynek Hermansky.
\newblock Perceptual linear predictive (plp) analysis of speech.
\newblock {\em the Journal of the Acoustical Society of America},
  87(4):1738--1752, 1990.

\bibitem[\protect\citeauthoryear{Hinton \bgroup \em et al.\egroup
  }{2006}]{hinton2006fast}
Geoffrey Hinton, Simon Osindero, and Yee-Whye Teh.
\newblock A fast learning algorithm for deep belief nets.
\newblock {\em Neural computation}, 18(7):1527--1554, 2006.

\bibitem[\protect\citeauthoryear{Hubel and Wiesel}{1962}]{hubel1962receptive}
David~H Hubel and Torsten~N Wiesel.
\newblock Receptive fields, binocular interaction and functional architecture
  in the cat's visual cortex.
\newblock {\em The Journal of physiology}, 160(1):106, 1962.

\bibitem[\protect\citeauthoryear{Kavukcuoglu \bgroup \em et al.\egroup
  }{2010}]{kavukcuoglu2010learning}
Koray Kavukcuoglu, Pierre Sermanet, Y-Lan Boureau, Karol Gregor, Micha{\"e}l
  Mathieu, and Yann~L Cun.
\newblock Learning convolutional feature hierarchies for visual recognition.
\newblock In {\em Advances in neural information processing systems}, pages
  1090--1098, 2010.

\bibitem[\protect\citeauthoryear{Keogh and Pazzani}{2000}]{keogh2000scaling}
Eamonn~J Keogh and Michael~J Pazzani.
\newblock Scaling up dynamic time warping for datamining applications.
\newblock In {\em Proceedings of the sixth ACM SIGKDD international conference
  on Knowledge discovery and data mining}, pages 285--289. ACM, 2000.

\bibitem[\protect\citeauthoryear{Keogh \bgroup \em et al.\egroup
  }{2011}]{keogh2011ucr}
Eamonn Keogh, Xiaopeng Xi, Li~Wei, and Chotirat~Ann Ratanamahatana.
\newblock The ucr time series classification/clustering homepage.
\newblock {\em URL= http://www. cs. ucr. edu/\~{} eamonn/time\_series\_data},
  2011.

\bibitem[\protect\citeauthoryear{Krizhevsky \bgroup \em et al.\egroup
  }{2012}]{krizhevsky2012imagenet}
Alex Krizhevsky, Ilya Sutskever, and Geoffrey~E Hinton.
\newblock Imagenet classification with deep convolutional neural networks.
\newblock In {\em Advances in neural information processing systems}, pages
  1097--1105, 2012.

\bibitem[\protect\citeauthoryear{LeCun \bgroup \em et al.\egroup
  }{1998}]{lecun1998gradient}
Yann LeCun, L{\'e}on Bottou, Yoshua Bengio, and Patrick Haffner.
\newblock Gradient-based learning applied to document recognition.
\newblock {\em Proceedings of the IEEE}, 86(11):2278--2324, 1998.

\bibitem[\protect\citeauthoryear{Lin \bgroup \em et al.\egroup
  }{2012}]{lin2012rotation}
Jessica Lin, Rohan Khade, and Yuan Li.
\newblock Rotation-invariant similarity in time series using bag-of-patterns
  representation.
\newblock {\em Journal of Intelligent Information Systems}, 39(2):287--315,
  2012.

\bibitem[\protect\citeauthoryear{Ngiam \bgroup \em et al.\egroup
  }{2010}]{ngiam2010tiled}
Jiquan Ngiam, Zhenghao Chen, Daniel Chia, Pang~W Koh, Quoc~V Le, and Andrew~Y
  Ng.
\newblock Tiled convolutional neural networks.
\newblock In {\em Advances in Neural Information Processing Systems}, pages
  1279--1287, 2010.

\bibitem[\protect\citeauthoryear{Rakthanmanon and
  Keogh}{2013}]{rakthanmanon2013fast}
Thanawin Rakthanmanon and Eamonn Keogh.
\newblock Fast shapelets: A scalable algorithm for discovering time series
  shapelets.
\newblock In {\em Proceedings of the thirteenth SIAM conference on data mining
  (SDM)}. SIAM, 2013.

\bibitem[\protect\citeauthoryear{Senin and Malinchik}{2013}]{senin2013sax}
Pavel Senin and Sergey Malinchik.
\newblock Sax-vsm: Interpretable time series classification using sax and
  vector space model.
\newblock In {\em Data Mining (ICDM), 2013 IEEE 13th International Conference
  on}, pages 1175--1180. IEEE, 2013.

\bibitem[\protect\citeauthoryear{Silva \bgroup \em et al.\egroup
  }{2013}]{silva2013time}
Diego~F Silva, Vinicius Souza, MA~De, and Gustavo~EAPA Batista.
\newblock Time series classification using compression distance of recurrence
  plots.
\newblock In {\em Data Mining (ICDM), 2013 IEEE 13th International Conference
  on}, pages 687--696. IEEE, 2013.

\bibitem[\protect\citeauthoryear{Vincent \bgroup \em et al.\egroup
  }{2008}]{vincent2008extracting}
Pascal Vincent, Hugo Larochelle, Yoshua Bengio, and Pierre-Antoine Manzagol.
\newblock Extracting and composing robust features with denoising autoencoders.
\newblock In {\em Proceedings of the 25th international conference on Machine
  learning}, pages 1096--1103. ACM, 2008.

\end{thebibliography}
\end{document}